\documentclass{article}

\usepackage[final]{neurips_2019}

\usepackage[utf8]{inputenc} % allow utf-8 input
\usepackage[T1]{fontenc}    % use 8-bit T1 fonts
\usepackage{hyperref}
\hypersetup{
    colorlinks=true,
    citecolor=black,
    linkcolor=black,
    filecolor=magenta,      
    urlcolor=cyan,
}
\usepackage{url}            % simple URL typesetting
\usepackage{booktabs}       % professional-quality tables
\usepackage{amsfonts}       % blackboard math symbols
\usepackage{nicefrac}       % compact symbols for 1/2, etc.
\usepackage{microtype}      % microtypography

\usepackage{times}
\usepackage{latexsym}
\usepackage{xcolor}

\usepackage{caption} 
\usepackage{wrapfig}
\usepackage{subfig}
\usepackage{graphicx}
\usepackage{bbm}
\usepackage{sidecap}
\usepackage{amsmath}

\title{Interpreting and improving natural-language processing (in machines) with natural language-processing (in the brain)}

\author{Mariya Toneva \\
  Neuroscience Institute \\
  Department of Machine Learning \\
  Carnegie Mellon University \\
  {\tt mariya@cmu.edu} \\\And
  Leila Wehbe \\
  Neuroscience Institute \\
  Department of Machine Learning \\
  Carnegie Mellon University \\
  {\tt lwehbe@cmu.edu} \\}

\date{}

\newcommand\blfootnote[1]{%
  \begingroup
  \renewcommand\thefootnote{}\footnote{#1}%
  \addtocounter{footnote}{-1}%
  \endgroup
}

\begin{document}

\maketitle

\begin{abstract}
Neural networks models for NLP are typically implemented without the explicit encoding of language rules and yet they are able to break one performance record after another.  This has generated a lot of research interest in interpreting the representations learned by these networks. We propose here a novel interpretation approach that relies on the only processing system we have that does understand language: the human brain.
We use brain imaging recordings of subjects reading complex natural text to interpret word and sequence embeddings from $4$ recent NLP models - ELMo, USE, BERT and Transformer-XL. We study how their representations differ across layer depth, context length, and attention type. Our results reveal differences in the context-related representations across these models. Further, in the transformer models, we find an interaction between layer depth and context length, and between layer depth and attention type. We finally hypothesize that altering BERT to better align with brain recordings would enable it to also better understand language. Probing the altered BERT using syntactic NLP tasks reveals that the model with increased brain-alignment outperforms the original model. Cognitive neuroscientists have already begun using NLP networks to study the brain, and this work closes the loop to allow the interaction between NLP and cognitive neuroscience to be a true cross-pollination.
\end{abstract}

\blfootnote{\\ Code available at \href{https://github.com/mtoneva/brain_language_NLP}{\texttt{https://github.com/mtoneva/brain\_language\_nlp}}}

\section{Introduction}

The large success of deep neural networks in NLP is perplexing when considering that unlike most other NLP approaches, neural networks are typically not informed by explicit language rules. Yet, neural networks are constantly breaking records in various NLP tasks from machine translation to sentiment analysis. Even more interestingly, it has been shown that word embeddings and language models trained on a large generic corpus and then optimized for downstream NLP tasks produce even better results than training the entire model only to solve this one task \citep{peters2018deep,howard2018universal,devlin2018bert}. These models seem to capture something generic about language. What representations do these models capture of their language input?

\begin{figure*}[th]
\centering
\includegraphics[width=\linewidth]{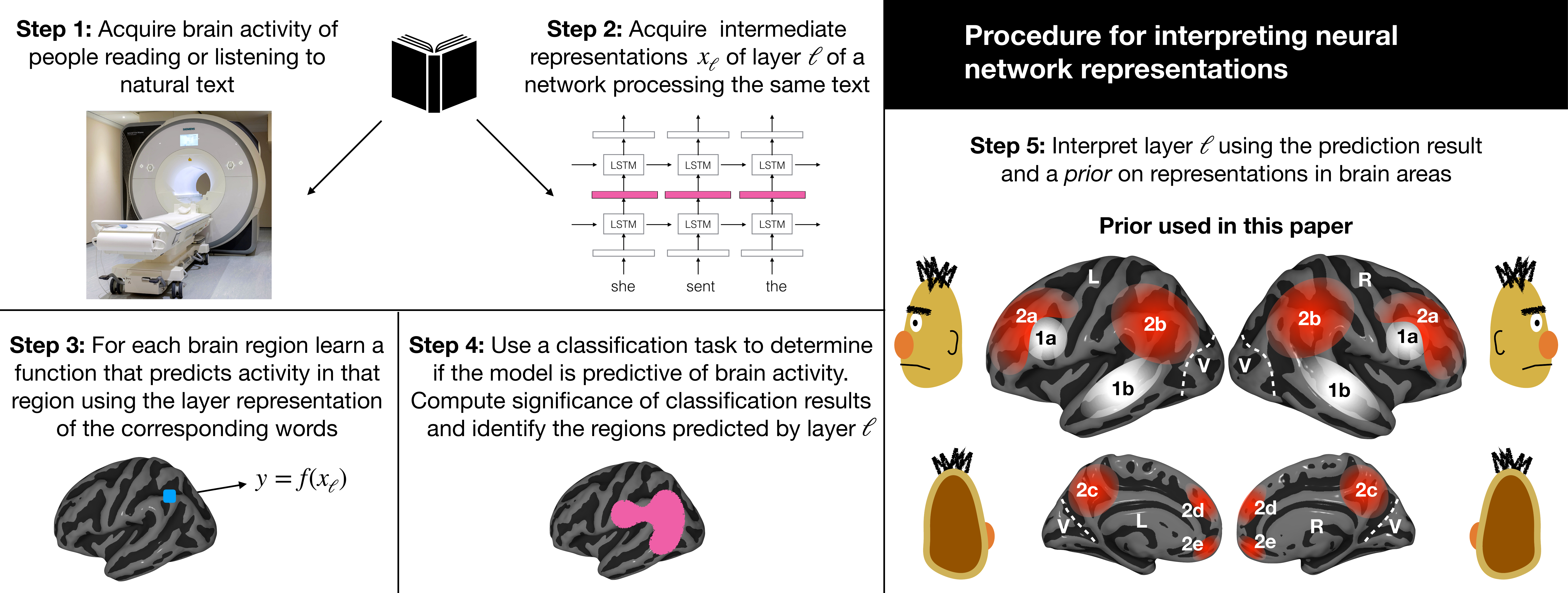}
\caption{Diagram of approach and prior on brain function. 
The prior was constructed using  the results of \citet{lerner2011topographic}: regions in group~1 (white) process information related to isolated words and word sequences while group~2 (red) process only information related to word sequences (see Section \ref{section:prior}). V indicates visual cortex. The drawing indicates the views of the brain with respect to the head.  See supplementary materials for names of brain areas and  full description of the methods.} 
\label{fig:priors}
\end{figure*}

Different approaches have been proposed to probe the representations in the network layers through NLP tasks designed to detect specific linguistic information \citep{conneau2018you,zhu2018exploring,linzen2016assessing}. Other approaches have attempted to offer a more theoretical assessment of how recurrent networks propagate information, or what word embeddings can represent \citep{peng2018rational,chen2017recurrent,weiss2018practical}. Most of this work has been centered around understanding the properties of sequential models such as LSTMs and RNNs, with considerably less work focused on non-sequential models such as transformers.

Using specific NLP tasks, word annotations or behavioral measures to detect if a type of information is present in a network-derived representation (such as a word embedding of an LSTM or a state vector of a transformer) can be informative. However, complex and arguably more interesting aspects of language, such as high level meaning, are difficult to capture in an NLP task or in behavioral measures. We therefore propose a novel approach for interpreting neural networks that relies on the only processing system we have that does understand language: the human brain. %While it can be hard to produce behavioral measures of , 
Indeed, the brain does represent complex linguistic information
while processing language, and we can use brain activity recordings as a proxy for these representations. We can then relate the brain representations with neural network representations by learning a mapping from the latter to the former. We refer to this analysis as aligning the neural network representations with brain activity. 

\subsection{Proposed approach}

We propose to look at brain activity of subjects reading naturalistic text as a source of additional information for interpreting neural networks. We use fMRI (functional Magnetic Resonance Imaging) and Magnetoencephalography (MEG) recordings of the brain activity of these subjects as they are presented text one word at a time.
We present the same text to the NLP model we would like to investigate and extract representations from the intermediate layers of the network, given this text. We then learn an alignment between these extracted representations and the brain recordings corresponding to the same words to offer an evaluation of the information contained in the network representations. Evaluating neural network representations with brain activity is a departure from existing studies that go the other way, using such an alignment to instead evaluate brain representations  \citep{hp_meg,frank2015erp,hale2018finding,jain2018incorporating}. 

To align a layer $\ell$ representation with brain activity, we first learn a model that predicts the fMRI or MEG activity in every region of the brain (fig. \ref{fig:priors}). We determine the regions where this model is predictive of brain activity using a classification task followed by a significance test. If a layer representation can accurately predict the activity in a brain region $r$, then we conclude that the layer shares information with brain region $r$. We can thus make conclusions about the representation in layer $\ell$ based on our prior knowledge of region $r$. 

Brain recordings have inherent, meaningful structure that is absent in network-derived representations.
In the brain, different processes are assigned to specific locations as has been revealed by a large array of fMRI experiments. These processes have specific latencies and follow a certain order, which has been revealed by electrophysiology methods such as MEG.
In contrast to the brain, a network-derived representation might encode information that is related to multiple of these processes without a specific organization. 
When we align that specific network representation with fMRI and MEG data, the result will be a decomposition of the representation into parts that correspond to different processes and should therefore be more interpretable. %that might be more or less interpretable.
We can think of alignment with brain activity as a ``demultiplexer" in which a single input (the network-derived representation) is decomposed into multiple outputs (relationship with different brain processes).

There doesn't yet exist a unique theory of how the brain processes language that  researchers agree upon \citep{hickok2007cortical, friederici2011brain, hagoort2003brain}. Because we don't know which of the existing theories are correct, %and because most of these theories are based on experiments with controlled language stimulus, 
we abandon the theory-based approach and adopt a fully data-driven approach. We focus on results from experiments that use naturalistic stimuli to derive our priors on  the function of specific brain areas during language processing.  These experiments have found that a set of regions in the temporo-parietal and frontal cortices are activated in language processing \citep{lerner2011topographic, hp_fmri, huth2016natural, blank2017domain} and are collectively referred to as the language network \citep{fedorenko2014reworking}. 
Using the results of \citet{lerner2011topographic} we subdivide this network into two groups of areas: group~1 is consistently activated across subjects when they listen to disconnected words or to complex fragments like sentences or paragraphs and group~2 is consistently activated only when they listen to complex fragments. We will use group~1 as our prior on brain areas that process information at the level of both short-range context (isolated words) and long-range context (multi-word composition), and group~2 as a prior on areas that process long-range context only. Fig.~\ref{fig:priors} shows a simple approximation of these areas on the Montreal Neurological Institute (MNI) template. Inspection of the results of \citet{jain2018incorporating} shows they corroborate the division of language areas into group~1 and group~2. Because our prior relies on experimental results and not theories of brain function, it is data-driven. \label{section:prior}

We use this setup to investigate a series of questions about the information represented in different layers of neural network models. We explore four recent models: {ELMo}, a language model by \citet{peters2018deep}, {BERT}, a transformer by \citet{devlin2018bert}, {USE} (Universal Sentence Encoder), a sentence encoder by \citet{cer2018universal}, and  {T-XL} (Transformer-XL), a transformer that includes a recurrence mechanism by \citet{dai2019transformer}. We investigate multiple questions about these networks. Is word-level specific information represented only at input layers? Does this differ across recurrent models, transformers and other sentence embedding methods? How many layers do we need to represent a specific length of context? Is attention affecting long range or short range context?

\paragraph{Intricacies}

As a disclaimer, we warn the reader that one should be careful while dealing with brain activity. Say a researcher runs a task $T$ in fMRI (e.g. counting objects on the screen) and finds it activates region $R$, which is shown in another experiment to also be active during process $P$ (e.g. internal speech). It is seductive to then infer that process $P$ is involved during task $T$. This ``reverse inference" can lead to erroneous conclusions, as region $R$ can be involved in more than one task \citep{poldrack2006can}. To avoid
this trap, we only interpret alignment between network-derived representations and brain regions if (1) the function of the region is well studied and we have some confidence on its function during a task similar to ours (e.g. the primary visual cortex processing letters on the screen or group 2 processing long range context) or (2) 
we show a brain region has overlap in the variance explained by the network-derived layer and by a specific process, in the same experiment.
We further take sound measures for reporting results: we cross-validate our models and report results on unseen test sets. Another possible fallacy is to directly compare the performance of layers from different networks and conclude that one network performs better than the other: information is likely organized differently across networks and such comparisons are misleading. Instead we only perform controlled experiments where we look at one network and vary one parameter at a time, such as context length, layer depth  or attention type.

\subsection{Contributions}

\begin{enumerate}
    \item We present a new method to interpret network representations and a proof of concept for it.
    \item We use our method to analyze and provide hypotheses about {ELMo}, {BERT}, {USE} and {T-XL}.
    \item We find the middle layers of transformers are better at predicting brain activity than other layers. We find that {T-XL}'s performance doesn't degrade as context is increased, unlike the other models'. We find that using uniform attention in early layers of BERT (removing the pretrained attention on the previous layer) leads to better prediction of brain activity.
    \item We show that when BERT is altered to better align with brain recordings (by removing the pretrained attention in the shallow layers), it is also able to perform better at NLP tasks that probe its syntactic understanding \citep{marvin2018targeted}.
    This result shows a transfer of knowledge from the brain to NLP tasks and validates our approach. 
\end{enumerate}

\section{Related work on brains and language}

 Most work investigating language in the brain has been done in a controlled experiment setup where two conditions are contrasted \citep{friederici2011brain}. These conditions  typically vary in complexity (simple vs. complex sentences), vary in the presence or absence of a linguistic property (sentences vs. lists of words) or vary in the presence or absence of incongruities (e.g. semantic surprisal) \citep{friederici2011brain}. A few  researchers instead use naturalistic stimulus such as stories \citep{brennan2010syntactic,lerner2011topographic,speer2009reading,hp_fmri,huth2016natural,blank2017domain}. Some use predictive models of brain activity as  a function of multi-dimensional features spaces describing the different properties of the stimulus \citep{hp_fmri,huth2016natural}. 

A few previous works have used neural network representations as a source of feature spaces to model brain activity. \citet{hp_fmri} aligned the MEG brain activity we use here with a Recurrent Neural Network (RNN), trained on an online archive of Harry Potter Fan Fiction. The authors  aligned brain activity with the context vector and the word embedding, allowing them to trace sentence comprehension at a word-by-word level. \cite{jain2018incorporating} aligned layers from a Long Short-Term Memory (LSTM) model to fMRI recordings of subjects listening to stories to differentiate between the amount of context maintained by each brain region.  Other approaches rely on computing surprisal or cognitive load metrics using neural networks to identify processing effort in the brain, instead of aligning entire representations \citep{frank2015erp,hale2018finding}. 

There is little prior work that evaluates or improves NLP models through brain recordings. \citet{sogaard2016evaluating} proposes to evaluate whether a word embedding contains cognition-relevant semantics by measuring how well they predict eye tracking data and fMRI recordings. \citet{fyshe2014interpretable} build a non-negative sparse embedding for individual words by constraining the embedding to also predict brain activity well and show that the new embeddings better align with behavioral measures of semantics.

\section{Approach}

\paragraph{Network-derived Representations}

The approach we propose in this paper is general and can be applied to a wide variety of current NLP models. We present four case-studies of recent models that have very good performance on downstream tasks:
{ELMO}, {BERT}, {USE} and {T-XL}.

\begin{itemize}
\item{ELMo} is a bidirectional language model that incorporates multiple layers of LSTMs. It can be used to derive contextualized embeddings 
by concatenating the LSTM output layers at that word with its non-contextualized embedding. We use a pretrained version of {ELMo} with $2$ LSTM layers provided by \citet{Gardner2017AllenNLP}.
\item{BERT} is a bidirectional model of stacked transformers that is trained to predict whether a given sentence follows the current sentence, in addition to predicting a number of input words that have been masked ~\citep{devlin2018bert}. Upon release, this recent model achieved state of the art across a large array of NLP tasks, ranging from question answering to named entity recognition. We use a pretrained model provided by Hugging Face~\footnote{\label{note1}https://github.com/huggingface/pytorch-pretrained-BERT/}.
We investigate the base {BERT} model, which has $12$ layers, $12$ attention heads, and $768$ hidden units. 
\item{USE} is a method of encoding sentences into an embedding \citep{cer2018universal} using a task similar to Skip-thought~\citep{kiros2015skip}. {USE} is able to produce embeddings in the same space for single words and passages of text of different lengths.
We use a version of {USE} from tensorflow hub trained with  a deep averaging network~\footnote{https://tfhub.dev/google/universal-sentence-encoder/2} that has $512$ dimensions. 
\item{T-XL} incorporates segment level recurrence into a transformer with the goal of capturing longer context than either recurrent networks or usual transformers \citep{dai2019transformer}. We use a pretrained model provided by Hugging Face$^{\ref{note1}}$, with $19$ layers and $1024$ hidden units.

\end{itemize}

We investigate how the representations of all four networks change as we provide varying lengths of context. We compute the representations $x_{\ell,k}$ in each available intermediate layer  ($\ell\in\{1,2\}$ for {ELMo}; $\ell \in\{1,..12\}$ for {BERT}; $\ell$ is the output embedding for {USE};  $\ell\in\{1,..19\}$ for {T-XL}). We compute $x_{l,k}$ for word $w_n$ by passing the most recent $k$ words ($w_{n-k+1},..,w_{n}$) through the network.

\paragraph{fMRI and MEG data}
In this paper we use fMRI and MEG data which have complementary strengths. fMRI is sensitive to the change in oxygen level in the blood that is a consequence to neural activity, it has high spatial resolution (2-3mm) and low temporal resolution (multiple seconds). MEG measures the change in the magnetic field outside the skull due to neural activity, it has low spatial resolution (multiple cm) and high temporal resolution (up to 1KHz).  We use fMRI data published by \citet{hp_fmri}. 8 subjects read chapter 9 of \emph{Harry Potter and the Sorcerer's stone} \citet{rowling2012harry} which was presented one word at a time for a fixed duration of 0.5 seconds each, and 45 minutes of data were recorded. The fMRI sampling rate (TR) was 2 seconds. The same chapter was shown by \citet{hp_meg} to 3 subjects in MEG with the same rate of 0.5 seconds per word. Details about the data and preprocessing can be found in the supplementary materials. 

\begin{figure}
\centering
\includegraphics[width=0.8\linewidth]{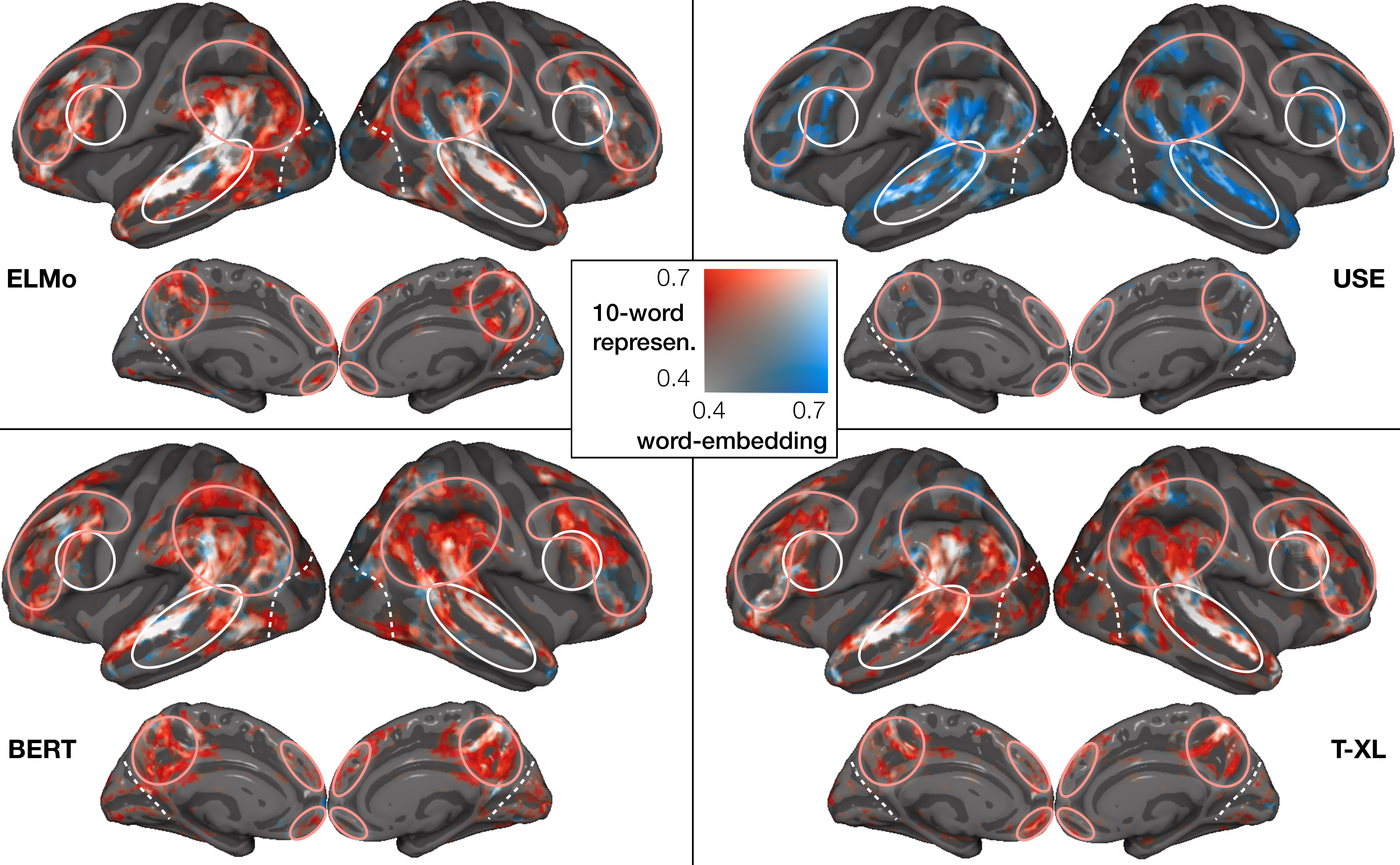}
\caption{Comparison between the prediction performance of two network representations from each model: a 10-word representation corresponding to the 10 most recent words shown to the participant (Red) and a word-embedding corresponding to the last word (Blue). Areas in white are well predicted from both representations. These results align to a fair extent with our prior: group~2 areas (red outlines) are mostly predicted by the longer context representations while areas 1b (lower white outlines) are predicted by both word-embeddings and longer context representations.  
}
\label{fig:four_brains}
\end{figure}

\begin{figure}
\centering
\includegraphics[width=0.9\linewidth]{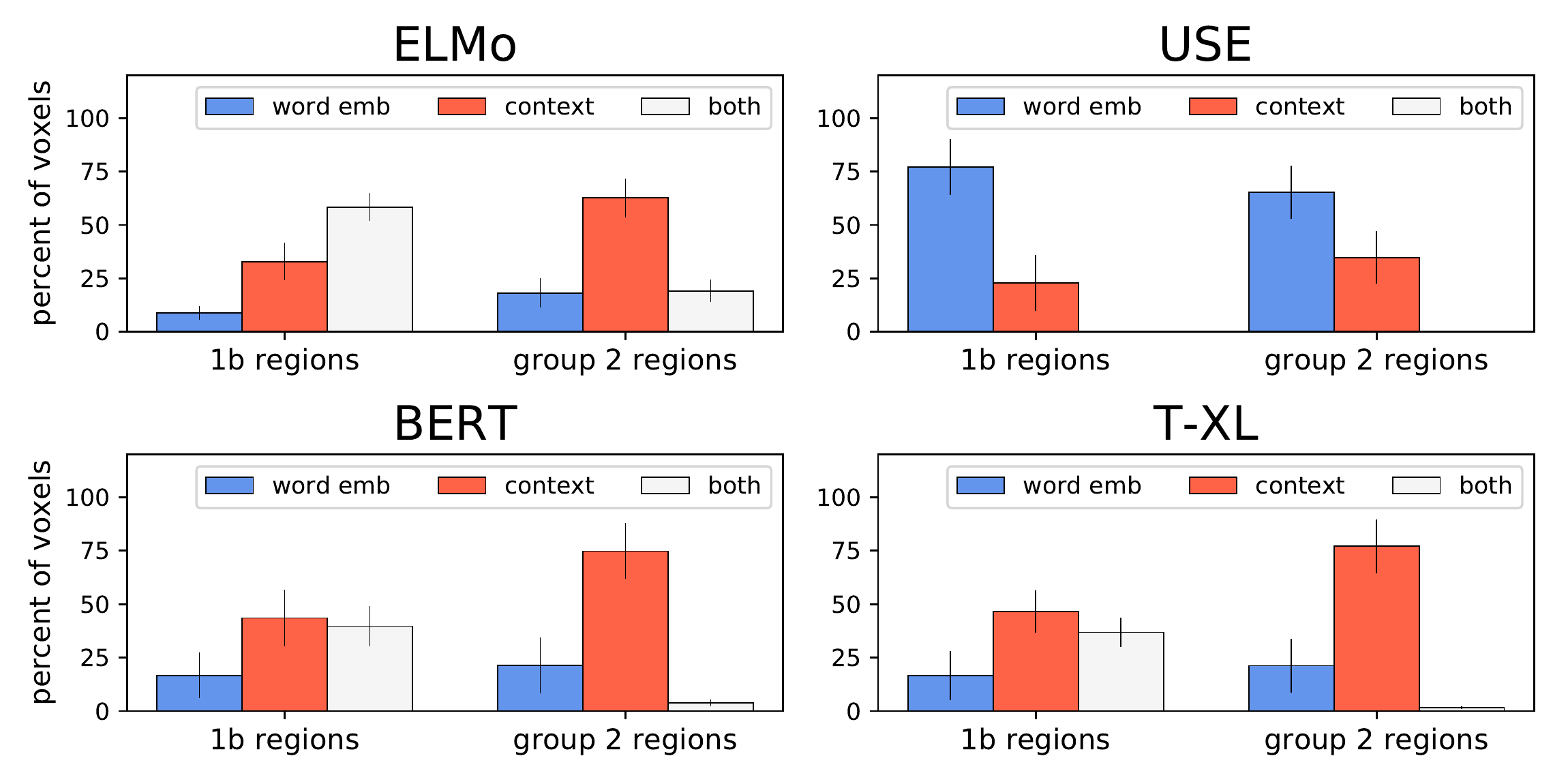}
\caption{Amount of group 1b regions and group 2 regions predicted well by each network-derived representation: a 10-word representation corresponding to the 10 most recent words shown to the participant (Red) and a word-embedding corresponding to the last word (Blue). White indicates that both representations predict the specified amount of the regions well (about 0.7 threshold). We present the mean and standard error of the percentage of explained voxels within the specified regions over all participants.
}
\label{fig:4barplots}

\end{figure}

\paragraph{Encoding models}
\label{subsec:encoding}
For each type of network-derived representation $x_{\ell,k}$, we estimate an encoding model that takes $x_{\ell,k}$ as input and predicts the brain recording associated with reading the same $k$ words that were used to derive $x_{\ell,k}$. We estimate a function $f$, such that $f(x_{l,k})=y$, where $y$ is the brain activity recorded with either MEG or fMRI. 
We follow previous work \citep{sudre2012tracking, hp_fmri, hp_meg, nishimoto2011reconstructing, huth2016natural} and model $f$ as a linear function, regularized by the ridge penalty. The model is trained via four-fold cross-validation and the regularization parameter is chosen via nested cross-validation.  

\paragraph{Evaluation of predictions}
We evaluate the predictions from each encoding model by using them in a classification task on held-out data, in the four-fold cross-validation setting. 
The classification task is to predict which of two sets of words was being read based on the respective feature representations of these words~\citep{mitchell2008predicting, hp_fmri,hp_meg}. This task is performed between sets of 20 consecutive TRs in fMRI (accounting for the slowness of the hemodynamic response), and sets of 20 randomly sampled words in MEG. The classification is repeated a large number of times and an average classification accuracy is obtained for each voxel in fMRI and for each sensor/timepoint in MEG. We refer to this accuracy of matching the predictions of an encoding model to the correct brain recordings as "prediction accuracy". The final fMRI results are reported on the MNI template, and we use pycortex to visualize them \cite{gao2015pycortex}.
See the supplementary materials for more details about our methods.

\paragraph{Proof of concept}
 Since MEG signals are faster than the rate of word presentation, they are more appropriate to study the components of word embeddings than the slow fMRI signals that cannot be attributed to individual words. We know that a word embedding learned from a text corpus is likely to contain information related to the number of letters and part of speech of a word. We show in section 4 of the supplementary materials
 that the number of letters of a word and its {ELMo} embedding predict a shared portion of brain activity early on (starting 100ms after word onset) in the back of the MEG helmet, over the visual cortex. Indeed, this region and latency are when we expect the visual information related to a word to be processed~\citep{sudre2012tracking}. Further, a word's part of speech and its {ELMo} embedding predict a shared portion of brain activity around 200ms after word onset in the left front of the MEG sensor. Indeed, we know from electrophysiology studies that part of speech violations incur a response around 200ms after word onset in the frontal lobe \citep{frank2015erp}. We conclude from these experiments that the {ELMo} embedding contains information about the number of letters and the part of speech of a word. Since we knew this from the onset, this experiment serves as a proof of concept for using our approach to interpret information in network representations.

\section{Interpreting long-range contextual representations}
\label{sec:context}

\begin{figure*}
  \centering
   \includegraphics[width=0.85\textwidth]{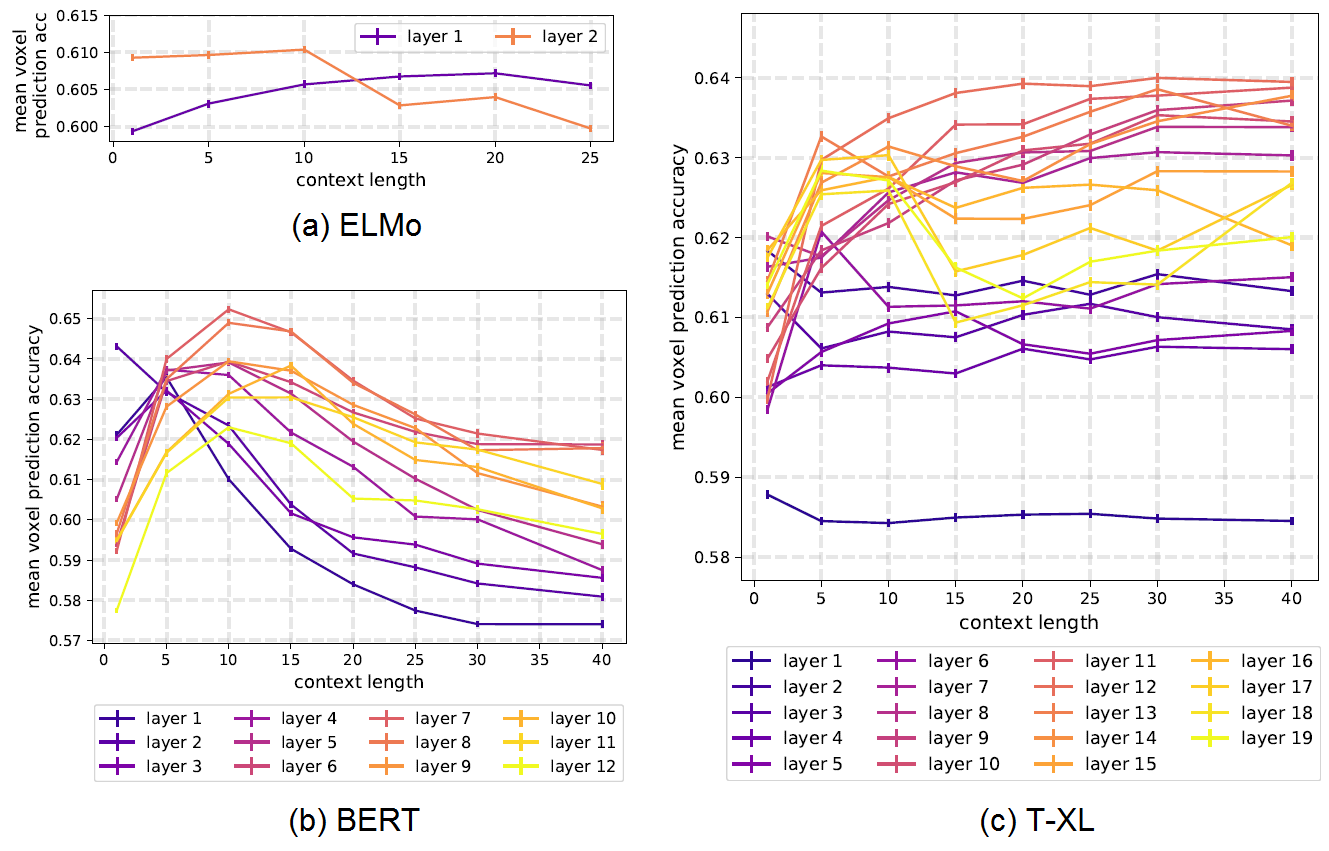}
\caption{Performance of encoding models for all hidden layers in {ELMo}, {BERT}, and {T-XL} as the amount of context provided to the network is increased. Transformer-XL is the only model that continues to increase performance as the context length is increased. In all networks, the middle layers perform the best for contexts longer than $15$ words. The deepest layers across all networks show a sharp increase in performance at short-range context (fewer than $10$ words), followed by a decrease in performance.}
\label{fig:layer_accs}
\end{figure*}

\paragraph{Integrated contextual information in {ELMo}, {BERT}, and {T-XL}} One question of interest in NLP is how successfully a model is able to integrate context into its representations. We investigate whether the four NLP models we consider are able to create an integrated representation of a text sequence by comparing the performance of encoding models trained with two kinds of representations: a token-level word-embedding corresponding to the most recent word token a participant was shown and a 10-word representation corresponding to the $10$ most recent words. For each of the models with multiple layers (all but {USE}), this 10-word representation was derived from a middle layer in the network (layer $1$ in {ELMo}, layer $7$ in {BERT}, and layer $11$ in {T-XL}). We present the qualitative comparisons across the four models in figure \ref{fig:four_brains}, where only significantly predicted voxels for each of the $8$ subjects were included with the false discovery rate controlled at level 0.05 (see section 3 of supplementary materials for more details). We provide a quantitative summary of the observed differences across models for the 1b regions and group 2 regions in Figure\ref{fig:4barplots}. We observe similarities in the word-embedding performances across all models, which all predict the brain activity in the left and right group 1b regions and to some extent in group 1a regions. We also observe differences in the longer context representations between {USE} and the rest of the models:
\begin{itemize}
    \item {ELMo}, {BERT}, and {T-XL} long context representations predict subsets of both group 1 regions and group 2 regions. 
     Most parts that are predicted by the word-embedding are also predicted by the long context representations (almost no blue voxels). We conclude that the long context representations most probably include information about the long range context and the very recent word embeddings. These results may be due to the fact that all these models are at least partially trained to predict a word at a given position. They must encode long range information and also local information that can predict the appropriate word.
    \item{USE} long context representations predict the activity in a much smaller subset of group 2 regions. The low performance of the {USE} vectors might be due to the deep averaging which might be composing words in a crude manner. The low performance in predicting group~1 regions is most probably because {USE} computes representations at a sentence level and does not have the option of retaining recent information like the other models. {USE} long context representations therefore only have long range information. 
\end{itemize}

\begin{figure}
\centering
\captionsetup{width=.40\textwidth}
\begin{minipage}{0.43\textwidth}
  \centering
  \includegraphics[width=\textwidth]{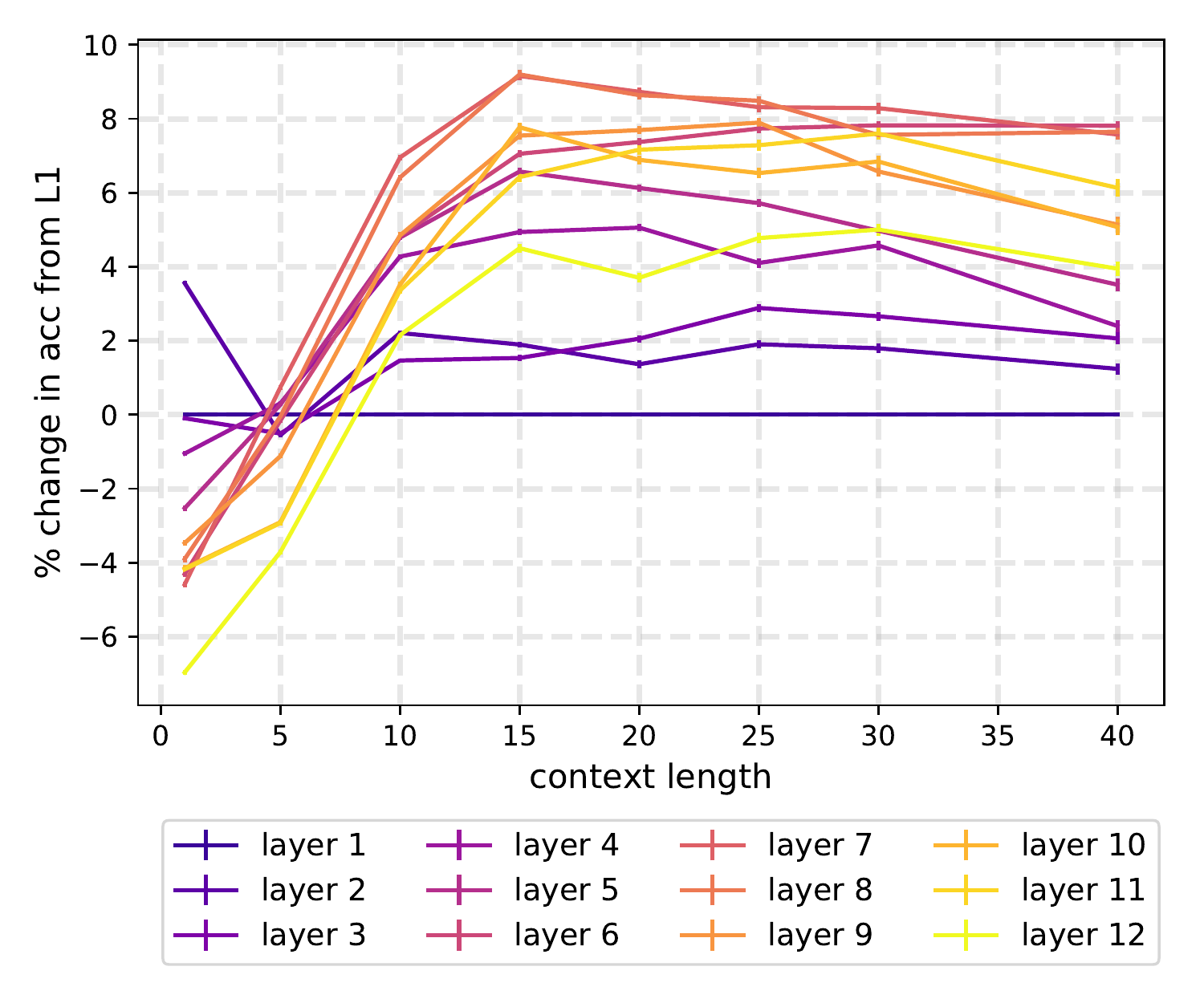}
  \captionof{figure}{Change in encoding model performance of {BERT} layers from the performance of the first layer. When we adjust for the performance of the first layer, the performance of the remaining layers resemble that of {T-XL} more closely, as shown in Figure~\ref{fig:layer_accs}.}
  \label{fig:l1_adjusted}
\end{minipage}%
\captionsetup{width=.43\textwidth}
\begin{minipage}{0.43\textwidth}
  \centering
  \includegraphics[width=\textwidth]{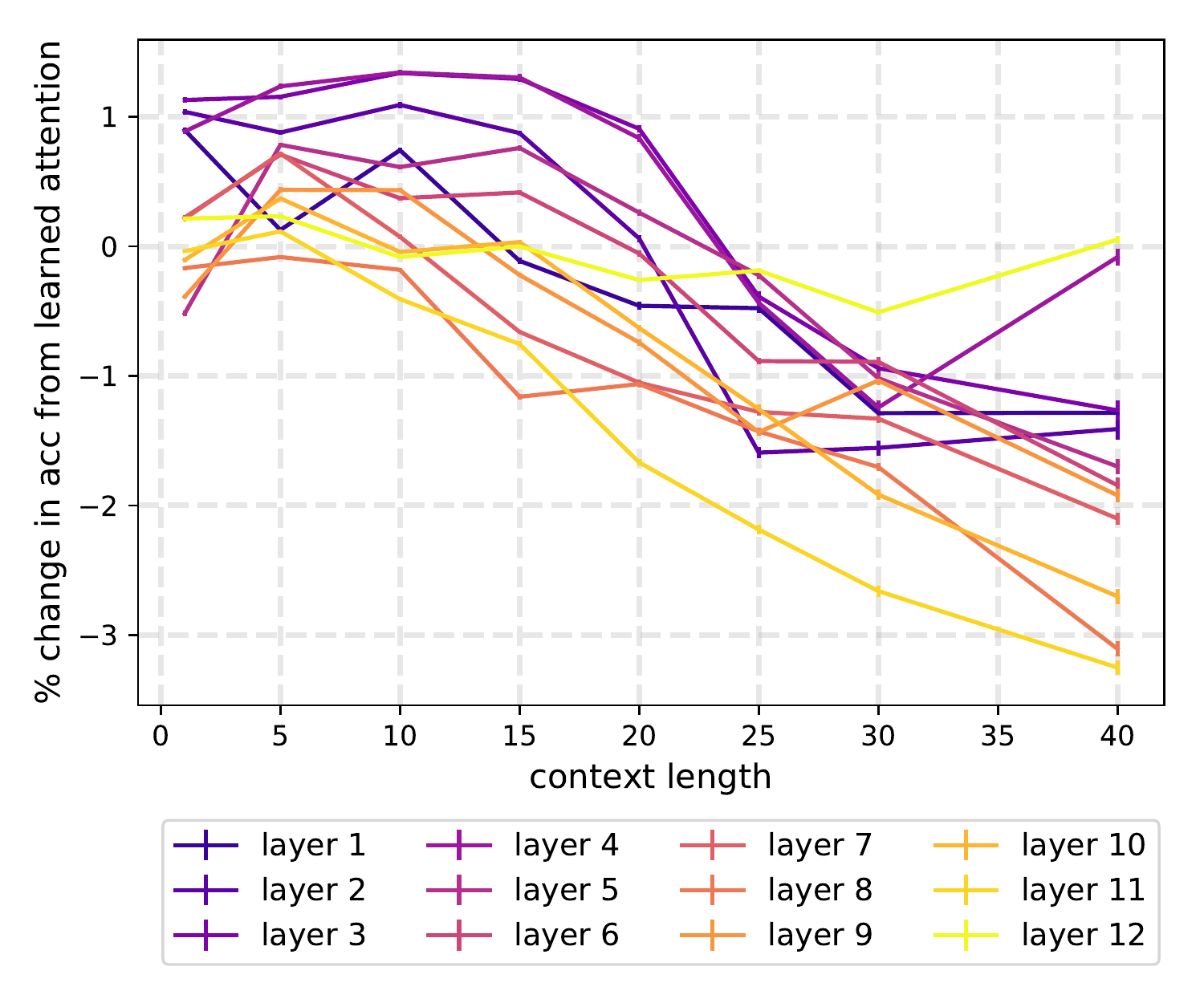}
  \captionof{figure}{Change in encoding model performance of {BERT} layer $l$ when the attention in layer $l$ is made uniform. The performance of deep layers, other than the output layer, is harmed by the change in attention. Shallow layers benefit from the uniform attention for context lengths up to $25$ words.}
  \label{fig:uni_attn}
\end{minipage}
\end{figure}

\paragraph{Relationship between layer depth and context length} 
We investigate how the performances of {ELMo}, {BERT}, and {T-XL}  change at different layers as they are provided varying size of contexts.
The results are shown in figure~\ref{fig:layer_accs}. We observe that in all networks, the middle layers perform the best for contexts longer than $15$ words. In addition, the deepest layers across all networks show a sharp increase in performance at short-range context (fewer than $10$ words), followed by a decrease in performance. We further observe that {T-XL} is the only model that continues to increase performance as the context length is increased. {T-XL} was designed to represent long range information better than a usual transformer and our results suggest that it does. Finally, we observe that layer $1$ in {BERT} behaves differently from the first layers in the other two networks. In figure~\ref{fig:l1_adjusted}, we show that when we instead examine the increase in performance of all subsequent layers from the performance of the first layer, the resulting context-layer relationships resemble the ones in {T-XL}. This suggests that {BERT} layer $1$ combines the information from the token-level embeddings in a way that limits the retention of longer context information in the layer $1$ representations.

\paragraph{Effect of attention on layer representation}
We further investigate the effect of attention across different layers by measuring the negative impact that removing its learned attention has on its brain prediction performance. Specifically we replaced the learned attention with uniform attention over the representations from the previous layer.
More concretely, to alter the attention pattern at a single layer in {BERT}, for each attention head $h_i = Attn_i(QW_i^Q,KW_i^K,VW_i^V)$, we replace the pretrained parameter matrices $W_i^Q$, $W_i^K$, and $W_i^V$ for this layer, such that the attention $Attn(Q,K,V)$, defined as $softmax({QK}/{\sqrt{d_k}})^TV$ ~\citep{vaswani2017attention}, yields equal probability over the values in value matrix $V$ (here $d_k$ denotes the dimensionality of the keys and queries). To this end, for a single layer, we replace $W_i^Q$ and $W_i^K$ with zero-filled matrices and $W_i^V$ with the identity matrix. We only alter a single layer at a time, while keeping all other parameters of the pretrained {BERT} fixed. In figure~\ref{fig:uni_attn}, we present the change in performance of each layer with uniform attention when compared to pretrained attention.  The performance of deep layers, other than the output layer, is harmed by the change in attention. However, surprisingly and against our expectations, shallow layers benefit from the uniform attention for context lengths up to $25$ words.

\vspace{-0.025in}
\section{Applying insight from brain interpretations to NLP tasks}
\vspace{-0.025in}

After observing that the layers in the first half of the base {BERT} model benefit from uniform attention for predicting brain activity, we test how the same alterations affect BERT's ability to predict language by testing its performance on natural language processing tasks. We evaluate on tasks that do not require fine-tuning beyond pretraining to ensure that there is an opportunity to transfer the insight from the brain interpretations of the pretrained {BERT} model. To this end, we evaluate on a range of syntactic tasks proposed by \citet{marvin2018targeted}, that have been previously used to quantify {BERT}'s syntactic capabilities~\citep{goldberg2019assessing}. These syntactic tasks measure subject-verb agreement in various types of sentences. They can be thought of as probe-tasks because they assess the ability of the network to perform syntax-related predictions without further fine-tuning.

We adopt the evaluation protocol of~\citet{goldberg2019assessing}, in which {BERT} is first fed a complete sentence where the single focus verb is masked (e.g.{\fontfamily{qcr}\selectfont [CLS] the game that the guard hates [MASK] bad .}), then the prediction for the masked position is obtained using the pretrained language-modeling head, and lastly the accuracy is obtained by comparing the scores for the original correct verb (e.g.{\fontfamily{qcr}\selectfont \quad is}) to the score for the incorrect verb (i.e. the verb that is wrongly numbered) (e.g.{\fontfamily{qcr}\selectfont \quad are}). We make the attention in layers $1$ through $6$ in base {BERT} uniform, a single layer at a time while keeping the remaining parameters fixed as described in Section~\ref{sec:context}, and evaluate on the $13$ tasks. We present the results of altering layers $1$,$2$, and $6$ in Table~\ref{table:nlp}. We observe that the altered models significantly outperform the pretrained model (`base') in $8$ of the $13$ tasks and achieve parity in $4$ of the remaining $5$ tasks (paired t-test, significance level 0.01, FDR controlled for multiple comparisons \citep{benjamini1995controlling}). Performance of altering layers 3-5 is similar and is presented in Supplementary Table 2. We contrast the performance of these layers with that of a model with uniform attention at layer $11$, which is the model that suffers the most from this change for predicting the brain activity as shown in Figure~\ref{fig:uni_attn}. We observe that this model also performs poorly on the NLP tasks as it performs on par or worse than the base model in $12$ of the $13$ tasks.

\begin{table}[t]
\centering
\begin{tabular}{|l|l|l|l|l|l|l|} 
\hline
condition & uni L1 & uni L2  & uni L6 & uni L11 & base & count\\
\hline
simple & \textbf{1.00} & \textbf{1.00} & \textbf{1.00} & 0.98 & \textbf{1.00} & 120\\
in a sentential complement & \textbf{0.83} & \textbf{0.83} & \textbf{0.83} & \textbf{0.83} & \textbf{0.83} & 1440\\
short VP coordination & 0.88 & 0.90 & \textbf{0.91} & 0.88 & 0.89 & 720\\
long VP coordination & 0.96 & 0.97 & \textbf{1.00**} & 0.96 & 0.98 & 400\\
across a prepositional phrase & 0.86 & \textbf{0.93**} & 0.88 & 0.82 & 0.85 & 19440\\
across a subject relative clause & 0.83 & 0.83 & \textbf{0.85**} & 0.83 & 0.84 & 9600\\
across an object relative clause & 0.87 & 0.91 & \textbf{0.92**} & 0.86 & 0.89 & 19680\\
across an object relative clause (no that) & \textbf{0.87} & 0.80 & \textbf{0.87} & 0.84 & 0.86 & 19680\\
in an object relative clause & \textbf{0.97**} & 0.95 & 0.91 & 0.93 & 0.95 & 15960\\
in an object relative clause (no that) & \textbf{0.83**} & 0.72 & 0.74 & 0.72 & 0.79 & 15960\\
reflexive anaphora: simple & 0.91 & 0.94 & \textbf{0.99**} & 0.95 & 0.94 & 280\\
reflexive anaphora: in a sent. complem. & 0.88 & 0.85 & 0.86 & 0.85 & \textbf{0.89} & 3360\\
reflexive anaphora: across rel. clause & 0.79 & \textbf{0.84**} & 0.79 & 0.76 & 0.80 & 22400\\
\hline
\end{tabular}
\vspace{0.1in}
\caption{Performance of models with altered attention on subject-verb agreement across various sentence types (tasks by \citet{marvin2018targeted}). Best performance per task is made bold, and marked with ** when difference from `base' performance is statistically significant. The altered models for the shallow layers significantly outperform the pretrained model (`base') in $8$ of the $13$ tasks and achieve parity in $4$ of the remaining $5$ tasks.}
\vspace{-0.1in}
\label{table:nlp}
\end{table}

\vspace{-0.025in}
\section{Discussion}
\vspace{-0.025in}

We introduced an approach to use brain activity recordings of subjects reading naturalistic text to interpret different  representations derived from neural networks. We used MEG to show that the (non-contextualized) word embedding of {ELMo} contains information about word length and part of speech as a proof of concept. We used fMRI to show that different network representation (for {ELMo}, {USE}, {BERT}, {T-XL}) encode information relevant to language processing at different context lengths. {USE} long-range context representations perform differently from the other model and do not also include short-range information. The transformer models ({BERT} and {T-XL}) both capture the most brain-relevant context information in their middle layers. {T-XL}, by combining both recurrent properties and transformer properties, has representations that don't degrade in performance when very long context is used, unlike purely recurrent models (e.g. ELMo) or transformers (e.g. BERT).

We found that uniform attention on the previous layer actually improved the brain prediction performance of the shallow layers (layers 1-6) over using learned attention. After this observation,  we tested how the same alterations affect BERT's ability to predict language by probing the altered BERT's representations using syntactic NLP tasks. We observed that the altered BERT performs better on the majority of the tasks. This result suggests that altering an NLP model to better align with brain recordings of people processing language may lead to better language understanding by the NLP model.

\vspace{-0.05in}
\paragraph{Future work}
We hope that as naturalistic brain experiments become more popular and data more widely shared, aligning brain activity with neural network will become a research area.
Our next steps are to expand the analysis using MEG to uncover new aspects of word-embeddings and to derive more informative fMRI brain priors that contain specific conceptual information that is linked to brain areas, and use them to study the high level semantic information in network representations. 

\subsubsection*{Acknowledgments}
We thank Tom Mitchell for valuable discussions. We thank the National Science Foundation for supporting this work through the Graduate Research Fellowship under Grant No. DGE1745016, and Google for supporting this work through the Google Faculty Award.

\small

\newcommand\BibTeX{B\textsc{ib}\TeX}

\bibliography{neurips2019}
\bibliographystyle{natbib}

\newpage

\section{Supplementary Materials}

\vspace{0.3in}

\section{Brain areas included in prior}

\begin{table}[h]
\centering
\begin{tabular}{|l|l|}
\hline
1a &   Inferior Frontal Gyrus   \\
\hline
1b &   Middle/Superior Temporal    \\
\hline
2a &  Lateral Middle/Superior Frontal    \\
\hline
2b & Supramarginal Gyrus / Posterior Superior Temporal / Angular Gyrus \\ 
\hline
2c & Precuneus \\
\hline
2d & Medial Superior Frontal  \\ 
\hline
2e &  Medial Orbito-Frontal\\
\hline
\end{tabular}
\vspace{0.1in}
\caption{Name of regions of interest in fig. 1 of main manuscript. Regions were approximated from the results of   \citep{lerner2011topographic}.}
\end{table}

\section{Data Preprocessing}

We use fMRI data of 8 subjects reading chapter 9 of \emph{Harry Potter and the Sorcerer's Stone} \citep{rowling2012harry}, collected and made available online by \citet{hp_fmri}\footnote{http://www.cs.cmu.edu/afs/cs/project/theo-73/www/plosone/}. Words were presented one at a time at a rate of 0.5s each. fMRI data was acquired at a rate of 2s per image, i.e. the repetition time (TR) is 2s. The images were comprised of $3\times3\times3mm$ voxels. The data for each subject was slice-time and motion corrected using SPM8 \citep{kay2008identifying}, then detrended and smoothed with a 3mm full-width-half-max kernel. The brain surface of each subject was reconstructed using Freesurfer \citep{fischl2012freesurfer}, and a grey matter mask was obtained. The Pycortex software \citep{gao2015pycortex} was used to handle and plot the data. For each subject, 25000-31000 cortical voxels were kept. 

The same paradigm was recorded for 3 subjects using MEG by the authors of \cite{hp_meg} and shared upon our request. This data was recorded at 306 sensors organized in 102 locations around the head. MEG records the change in magnetic field due to neuronal activity and the data we used was sampled
at 1kHz, then preprocessed using the  Signal Space Separation method (SSS) \citep{taulu2004suppression} and its temporal extension (tSSS) \citep{taulu2006spatiotemporal}.
The signal in every sensor was downsampled into 25ms non-overlapping time bins. For each of the 5176 word in the chapter, we therefore obtained a recording for 306 sensors at 20 time points after word onset (since each word was presented for 500ms).

\section{Encoding Models}

\subsection{fMRI}
\label{subsec:supp_fmri}
Ridge regularization is used to estimate the parameters of a linear model that predicts the brain activity $y^i$ in every fMRI voxel $i$ as a linear combination of a particular layer representation $x^\ell$. For each output dimension (voxel), the Ridge regularization parameter is chosen independently by nested cross-validation. We use Ridge regression because of its computational efficiency and because of the results of \citet{shrinkage} showing that for fMRI data, as long as proper regularization is used and the regularization parameter is chosen by cross-validation for each voxel independently, different regularization techniques lead to similar results. Indeed, Ridge regression is indeed a common regularization technique used for building predictive fMRI  \citep{mitchell2008predicting,nishimoto2011reconstructing,hp_fmri,huth2016natural}.

For every voxel $i$, a model is fit to predict the signals $y^i~=~[y^i_1,y^i_2,\dots,y^i_n]$, where $n$ is the number of time points, as a function of the representation derived from layer $\ell$ of a network. The words presented to the participants are first grouped by the TR interval in which they were presented. Then, the features of layer $\ell$ of the words in every group are averaged to form a sequence of features $x^\ell = [x^\ell_1, x^\ell_2, \dots, x^\ell_n]$ which are aligned with the brain signals. The models are trained to predict the signal at time $t$, $y_t$, using the concatenated vector $z^\ell_t$ formed of $[x^\ell_{t-1}, x^\ell_{t-2}, x^\ell_{t-3},x^\ell_{t-4}]$. The features of the words presented in the previous volumes are included in order to account for the lag in the hemodynamic response that fMRI records. Indeed, the response measured by fMRI is an indirect consequence  of  brain activity that peaks about 6 seconds after stimulus onset, and the solution of expressing brain activity as a function of the features of the preceding time points is a common solution for building predictive models \citep{nishimoto2011reconstructing,hp_fmri,huth2016natural}.

For each given subject and each layer $\ell$, we perform a cross-validation procedure to estimate how predictive that layer is of brain activity in each voxel $i$. For each fold:
\begin{itemize}
    \item The fMRI data $Y$ and feature matrix $Z^\ell = z^\ell_1,z^\ell_2,\ldots z^\ell_n$ are split into corresponding train  and validation matrices and these matrices are individually normalized (to get a mean of 0 and standard deviation of 1 for each voxel across time), ending with train matrices $Y^R$ and $Z^{R,\ell}$ and validation matrices $Y^V$ and $Z^{V,\ell}$.
    \item Using the train fold, a model $w^{i,\ell}$ is estimated as:
    
    \begin{align*}
        \arg\min_{w^{i,\ell}} ||y^{R,i}-Z^{R,\ell} w^{i,\ell}|_2^2 + \lambda^i||w^{i,\ell}||_2^2
    \end{align*}
    
    A ten-fold nested cross-validation procedure is first used to identify the best $\lambda^i$ for every voxel $i$ that minimizes nested cross-validation error. $w^{i,\ell}$ is then estimated using  $\lambda^i $ on the entire training fold.
    \item The predictions for each voxel on the validation fold are obtained as $ p^\ell =  Z^{V,\ell}w^{i,\ell} $.
    \item A classification task is then performed to assess the prediction performance of the learned model. This classification task is based on searchlight classification \citep{kriegeskorte2006information}, in which a sliding window groups each voxel with its immediate neighbors in the 3D grid of voxels. We perform a more accurate searchlight analysis we refer to as cortical-searchlight. We are interested only in the grey matter voxels (which contain neurons) and these comprise the most external part of the brain: the cortical sheet. The cortical sheet of each subject is highly folded, and voxels that lie in a neighborhood on the sheet are not necessarily neighbors in the 3D grid of voxels.  Using the reconstructed cortical sheet of each subject, we estimate for each cortical voxel a surrounding neighborhood by including the voxels adjacent to it on the cortical sheet, and the voxels adjacent to those voxels. See figure \ref{suppl:searchlight}. We use for each voxel $i$ this neighborhood of  voxels $N^i$ with $|N^i| = k^i$ in a classification task. 
    
  \begin{figure*}[ht]
\centering
\includegraphics[width=0.4\linewidth]{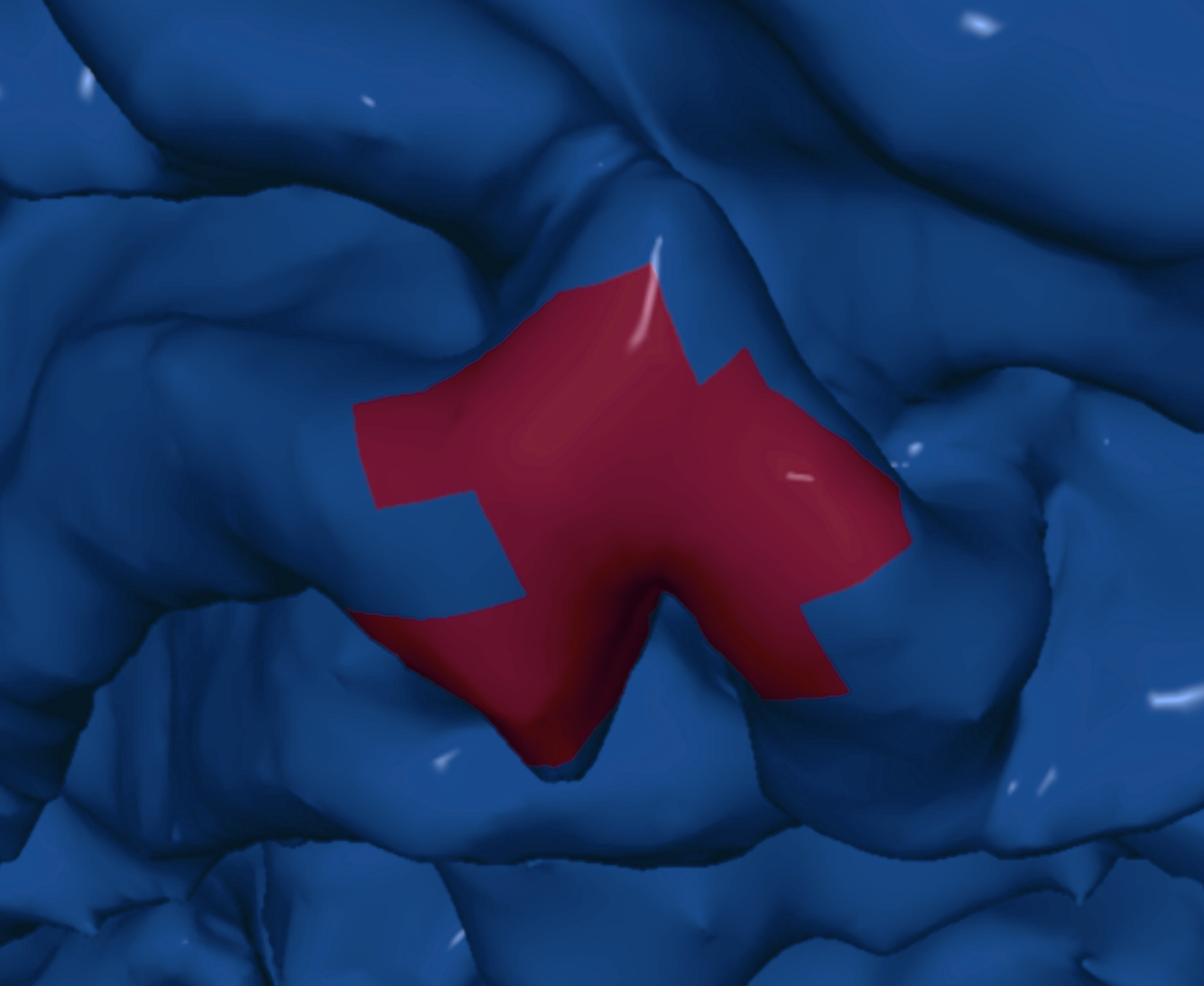}
\caption{Example neighborhood estimated using the cortical sheet and not the 3D grid of voxels.} 
\label{suppl:searchlight}
\end{figure*}

    \item For each voxel $i$, we use the signals predicted for layer $\ell$ to classify a contiguous chunk of real data of length 20TRs. Since fMRI data is noisy, performance using a single TR will be close to chance accuracy and will therefore have low power and will not be informative for our purpose. Indeed, for this reason most experiments using predictive fMRI models test them on a part of the experiment that is repeated multiple times \citep{kay2008identifying,nishimoto2011reconstructing,huth2016natural}. These repetitions are then averaged into one test set which is predicted, and this less noisy average leads to better prediction accuracy. The experiment we are using however doesn't have any repetitions and not specific test set, and therefore by raise the number of TRs and classify 20TRs at a time, we are able to improve the classification accuracy. \citet{hp_fmri} have shown that classification accuracy reaches a plateau after around 15 TRs and we pick 20TRs for good measure. The classification task takes an unlabeled chunk of real data of size $20\times k$ and two possible predicted data chunks of the same size, one being the predicted data corresponding to the same time, and another randomly chosen chunk. Euclidean distance is computed between the real chunk and the two predicted chunks, and the closest chunk is chosen. This is repeated a large number of times and average accuracy is computed at each voxel. 
\end{itemize}
The above steps are repeated for each of the four cross-validation folds and average accuracy is obtained for each voxel $i$ for layer $\ell$, for each subject.

We use a new empirical based method to compute statistical significance that relies on the distribution of average accuracies over a subject's brain to estimate the False Discovery Proportion (FDP). The voxel accuracies belong to two distributions: either the voxel has chance accuracy or the voxel is truly predicted by the corresponding layer $\ell$.
Average chance accuracy for our binary balanced task is 0.5, however the accuracies due to chance performance might have a varying distribution around 0.5. The accuracies above 0.5 are a mixture of predicted voxels and voxels with chance performance. We assume that chance performance is symmetrically distributed around 0.5, and we use the set of accuracies that are less than 0.5--which we consider to be in the chance distribution--to estimate the distribution of chance accuracies above 0.5.  We want to find a set of voxels where to reject the null hypothesis such that the FDP is $\leq0.05$. For that purpose we find the smallest margin $\delta, ~~ 0<\delta<0.5$ such that:

$$\widehat{\mathrm{FDP}} = \frac{1 + \# \{\mathrm{voxel} ~s.t.~ \mathrm{accuracy} \leq 0.5-\delta\}}{ 1\vee \# \{\mathrm{voxel} ~s.t.~ \mathrm{accuracy} \geq 0.5+\delta\}}\leq q$$

where $q = 0.05$, by starting at $\delta=0.001$ and increasing it in increments of 0.001, stopping when $\widehat{\mathrm{FDP}}\leq 0.05$ or the limit is reached. This approach is adapted from the Barber-Candès approach which has been proposed and analyzed by \cite{barber2015controlling,arias2017distribution,rabinovich2017optimal}, and shown to control the False Discovery Rate (FDR) at level $q$ when $\delta_{\textrm{final}}$ is chosen as a threshold.
We reject the null hypothesis for all voxels where the accuracy is $\ge0.5+\delta_{\textrm{final}}$.

To combine results across different subjects, we use pycortex \citep{gao2015pycortex} to transform each subject to the Montreal Neurological Institute (MNI) space, the most commonly used template space in fMRI. We can then average the results of different participants. 

\subsection{MEG}

MEG data is sampled faster than the rate of word presentation, so for each word, we have 20 times points recorded at 306 sensors.
Ridge regularization is similarly used to estimate the parameters of a linear model that predicts the brain activity $y^{i,\tau}$ in every MEG sensor $i$ at time $\tau$ after word onset. For each output dimension (sensor/time tuple $i,\tau$), the Ridge regularization parameter is chosen independently by nested cross-validation. 

For every sensor/time tuple $i,\tau$, a model is fit to predict the signals $y^{i,\tau}~=~[y^{i,\tau}_1,y^{i,\tau}_2,\dots,y^{i,\tau}_n]$, where $n$ is the number of words in the story, as a function of the representation derived from layer $\ell$ of a network. We use as input the word vector $x^\ell$ without the delays we used in fMRI because the MEG recordings capture instantaneous consequences of brain activity (change in the magnetic field). The models are trained to predict the signal at word $t$, $y^{i,\tau}_t$, using the  vector $x^\ell_t$.

For each each given subject and each layer $\ell$, we perform a cross-validation procedure to estimate how predictive that layer is of brain activity in each voxel $i$. For each fold:
\begin{itemize}
    \item The MEG data $Y$ and feature matrix $X^\ell = x^\ell_1,x^\ell_2,\ldots x^\ell_n$ are split into corresponding train  and validation matrices and these matrices are individually normalized (to get a mean of 0 and standard deviation of 1 for each voxel across time), ending with train matrices $Y^R$ and $X^{R,\ell}$ and validation matrices $Y^V$ and $Z^{V,\ell}$.
    \item Using the train fold, a model $w^{(i,\tau)\ell}$ is estimated as:
    
    \begin{align*}
        \arg\min_{w^{(i,\tau)\ell}} ||y^{(i,\tau),R}-X^{R,\ell} w^{(i,\tau)\ell}|_2^2 + \lambda^{(i,\tau)}||w^{(i,\tau)\ell}||_2^2
    \end{align*}
    
    A ten-fold nested cross-validation procedure is first used to identify the best $\lambda^{(i,\tau)}$ for every sensor, time-point tuple $(i,\tau)$ that minimizes nested cross-validation error. $w^{(i,\tau)\ell}$ is then estimated using  $\lambda^{(i,\tau)} $ on the entire training fold.
    \item The predictions for each sensor, time-point tuple $(i,\tau)$ on the validation fold are obtained as $ p^\ell =  X^{V,\ell}w^{(i,\tau)\ell} $.
    \item A classification task is then performed to assess the prediction performance of the learned model. This classification task also pools spatially: we use the 3 sensors at each location, pooling across all the subjects, ending up with 102 classifications at 20 time-points. By pooling the data in each sensor location across subjects, we increase the signal-to-noise ratio.
    \item For each sensor location $s$ and time-point $\tau$, we use the signals predicted from layer $\ell$ for the three sensors at time-point $\tau$ after word onset to classify a set of 20 words. Since MEG data is noisy, performance using a single word will be close to chance accuracy and will therefore have low power and will not be informative for our purpose. Indeed, for this reason most experiments using predictive MEG models test them on a part of the experiment that is repeated multiple times \citep{sudre2012tracking}. These repetitions are then averaged into one test set which is predicted, and this less noisy average leads to better prediction accuracy. The experiment we are using however doesn't have any repetitions and not specific test set, and therefore by raising the number of words and classify 20 words at a time, we are able to improve the classification accuracy. We use the value of 20 words from \citet{hp_meg}.
\end{itemize}
The above steps are repeated for each of the four cross-validation folds and average accuracy is obtained for each sensor location, time-point tuple $(s,\tau)$ for layer $\ell$, for each subject.

In our proof of concept experiment, we run an analysis in which we try to find, using the classification task outlined here, classification accuracy that is common both to a word embedding $\ell$ and to other features of a word such as a one-hot vector encoding its part of speech. This analysis is a proxy for finding the shared explained variance between the vectors, which we can call A and B. We concatenate A and B into a vector (representing $A\cup B$). We run the classification analysis using $A$, $B$ and ($A\cup B$). We then estimate the shared accuracy as: $A + B - A \cup B$.

\section{MEG results as proof of concept}

We use MEG to provide a proof of concept of our approach. We know that single word non-contextualized embeddings likely have information about the part-of-speech and the length of a word. We will show here how our approach can recover this information from brain activity as a proof-of-concept.  We use MEG to study word embeddings because unlike fMRI we can access the brain activity to reading a single word. 
%Following the novel finding that fMRI and MEG seem to relate to different types of network-derived representations, we further aim to investigate non-contextualized embeddings using MEG. 
We know from the Neuroscience literature that MEG activity can be related to the length of the current word~\cite{sudre2012tracking} and its part of speech~\cite{frank2015erp} at different times. We investigate whether word length and part-of-speech (POS) information is also present in the non-contextualized embedding by computing the shared performance ($A \cap B$) between the pairs of features ($A$ and $B$) as $A + B - A \cup B$ as explained in the previous section. 

\begin{figure*}[t]
\centering
\includegraphics[width=\linewidth]{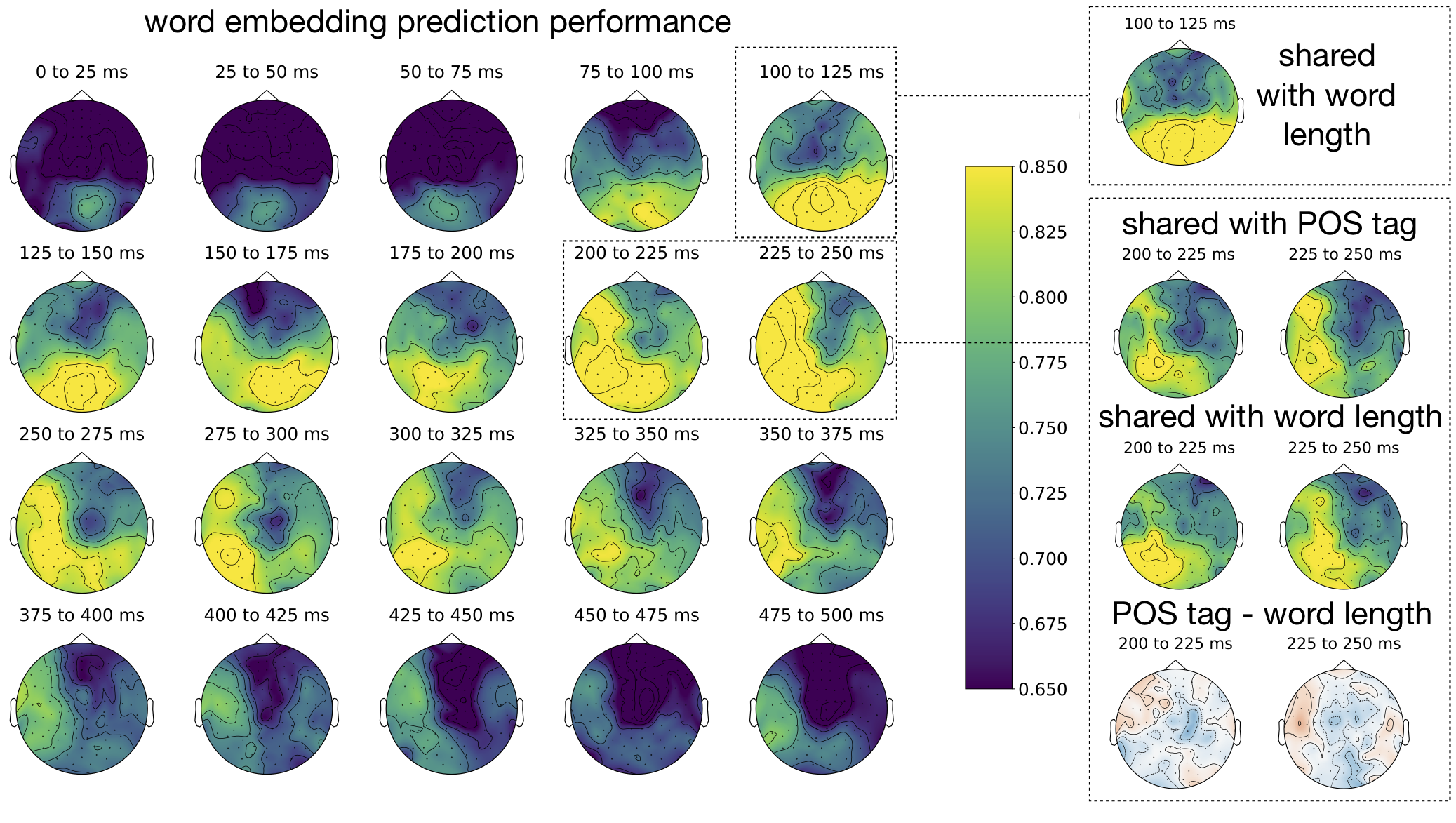}
\caption{Performance of {ELMo} current word embedding at predicting MEG activity at each sensor location and time point, compared with the performance shared with word length and Part-Of-Speech (POS) tags. Around 200-250ms, the word embedding predicts a part of the activity at the top of the helmet, and this is shared mostly with the POS tags and not with word length (see  bottom-right comparison). Indeed, we know from electrophysiology studies studies that POS violations incur a response around 200ms after word onset in the front of the brain   \cite{frank2015erp}, which aligns with our analysis. We hypothesize from these results that the word-embedding contains both word length and POS information.  }
\label{fig:meg_details}
\end{figure*}

We present the results in Figure~\ref{fig:meg_details}. The current word embedding is able to predict activity as the current word is being perceived starting at the back of the sensor helmet (generally on top of the visual cortex) around 100ms. This is when we expect the visual signal to start reaching the visual cortex. Indeed, we see that the word-embedding and the word length have overlap in the activity they predict in the visual cortex at that time. Gradually, the areas predicted by the word embedding move forward in the brain towards areas known to be involved in more high level aspects of reading. Around 200-250ms, we see the word embedding predicts a part of the activity at the top of the helmet, and this is shared mostly with the POS tags and not with word length (see  bottom-right comparison). Indeed, we know from electrophysiology studies  studies that POS violations incur a response around 200ms after word onset in the front of the brain   \cite{frank2015erp}, which aligns with our analysis. From these results we can hypothesize that the word-embedding contains both word length and POS information, as was expected.% This analysis therefore is a proof-of-concept of our method

\newpage
\section{Complete Attention Results}

\begin{table}[h]
\centering
\begin{tabular}{|l|l|l|l|l|l|l|l|l|}
\hline
condition & uni L1 & uni L2 & uni L3 & uni L4 & uni L5 & uni L6 & base & count\\ \hline
simple & 1.00 & 1.00 & 0.96 & 1.00 & 0.99 & 1.00 & 1.00 & 120\\
in a sentential complement & 0.83 & 0.83 & 0.83 & 0.83 & 0.84 & 0.83 & 0.83 & 1440\\
short VP coordination & 0.88 & 0.90 & 0.91 & 0.88 & 0.88 & 0.91 & 0.89 & 720\\
long VP coordination & 0.96 & 0.97 & 0.95 & 0.95 & 0.96 & 1.00 & 0.98 & 400\\
across a prepositional phrase & 0.86 & 0.93 & 0.88 & 0.86 & 0.80 & 0.88 & 0.85 & 19440\\
across a subject relative clause & 0.83 & 0.83 & 0.84 & 0.84 & 0.83 & 0.85 & 0.84 & 9600\\
across an object relative clause & 0.87 & 0.91 & 0.90 & 0.86 & 0.83 & 0.92 & 0.89 & 19680\\
across an object relative clause (no that) & 0.87 & 0.80 & 0.75 & 0.72 & 0.75 & 0.87 & 0.86 & 19680\\
in an object relative clause & 0.97 & 0.95 & 0.96 & 0.92 & 0.91 & 0.91 & 0.95 & 15960\\
in an object relative clause (no that) & 0.83 & 0.72 & 0.70 & 0.69 & 0.74 & 0.74 & 0.79 & 15960\\
reflexive anaphora: simple & 0.91 & 0.94 & 0.99 & 0.98 & 1.00 & 0.99 & 0.94 & 280\\
reflexive anaphora: in a sent. complem. & 0.88 & 0.85 & 0.88 & 0.87 & 0.86 & 0.86 & 0.89 & 3360\\
reflexive anaphora: across a rel. clause & 0.79 & 0.84 & 0.82 & 0.68 & 0.66 & 0.79 & 0.80 & 22400\\
\hline
\end{tabular}
\vspace{0.1in}
\caption{Performance of models with uniformly-altered attention in layers 1-6 in {BERT}
on a range of syntactic tasks by \citet{marvin2018targeted}. `Base' refers to pretrained BERT.}
\vspace{0.1in}
\label{table:supp_results}
\end{table}

\clearpage

\end{document}